\documentclass[letterpaper]{article} 
\usepackage{aaai25}  

\usepackage{graphicx}
\usepackage{multirow}
\usepackage{booktabs}
\usepackage{amsmath} 
\usepackage{makecell}

\usepackage{times}  
\usepackage{helvet}  
\usepackage{courier}  
\usepackage[hyphens]{url}  
\usepackage{graphicx} 
\usepackage{natbib}  
\usepackage{caption} 
\usepackage{algorithm}
\usepackage{algorithmic}

\usepackage{newfloat}
\usepackage{listings}
\DeclareCaptionStyle{ruled}{labelfont=normalfont,labelsep=colon,strut=off} 
\floatstyle{ruled}
\newfloat{listing}{tb}{lst}{}
\floatname{listing}{Listing}
\title{Mind the Uncertainty in Human Disagreement: Evaluating Discrepancies\\between Model Predictions and Human Responses in VQA}
\author {
    Jian Lan\textsuperscript{\rm 1,\rm 2},
    Diego Frassinelli \textsuperscript{\rm 1,\rm 2},
    Barbara Plank\textsuperscript{\rm 1, \rm 2}
}
\affiliations {
    \textsuperscript{\rm 1} MaiNLP, Center for Information and Language Processing, LMU Munich, Germany\\
    \textsuperscript{\rm 2}Munich Center for Machine Learning (MCML), Munich, Germany\\
    \text{\{}jianlan, frassinelli, bplank\text{\}}@cis.lmu.de
}

\usepackage{bibentry}

\begin{document}

\maketitle

\begin{abstract}
Large vision-language models frequently struggle to accurately predict responses provided by multiple human annotators, particularly when those responses exhibit human uncertainty. In this study, we focus on the Visual Question Answering (VQA) task, and we comprehensively evaluate how well the state-of-the-art vision-language models correlate with the distribution of human responses. To do so, we categorize our samples based on their levels (low, medium, high) of human uncertainty in disagreement (HUD) and employ not only accuracy but also three new human-correlated metrics in VQA, to investigate the impact of HUD. To better align models with humans, we also verify the effect of common calibration and human calibration \cite{stop}. Our results show that even BEiT3, currently the best model for this task, struggles to capture the multi-label distribution inherent in diverse human responses. Additionally, we observe that the commonly used accuracy-oriented calibration technique adversely affects BEiT3's ability to capture HUD, further widening the gap between model predictions and human distributions. In contrast, we show the benefits of calibrating models towards human distributions for VQA, better aligning model confidence with human uncertainty. Our findings highlight that for VQA, the consistent alignment between human responses and model predictions is understudied and should become the next crucial target of future studies.
\end{abstract}

\section{Introduction}

\begin{figure}[t!]
    \centering
    \includegraphics[width=\linewidth]{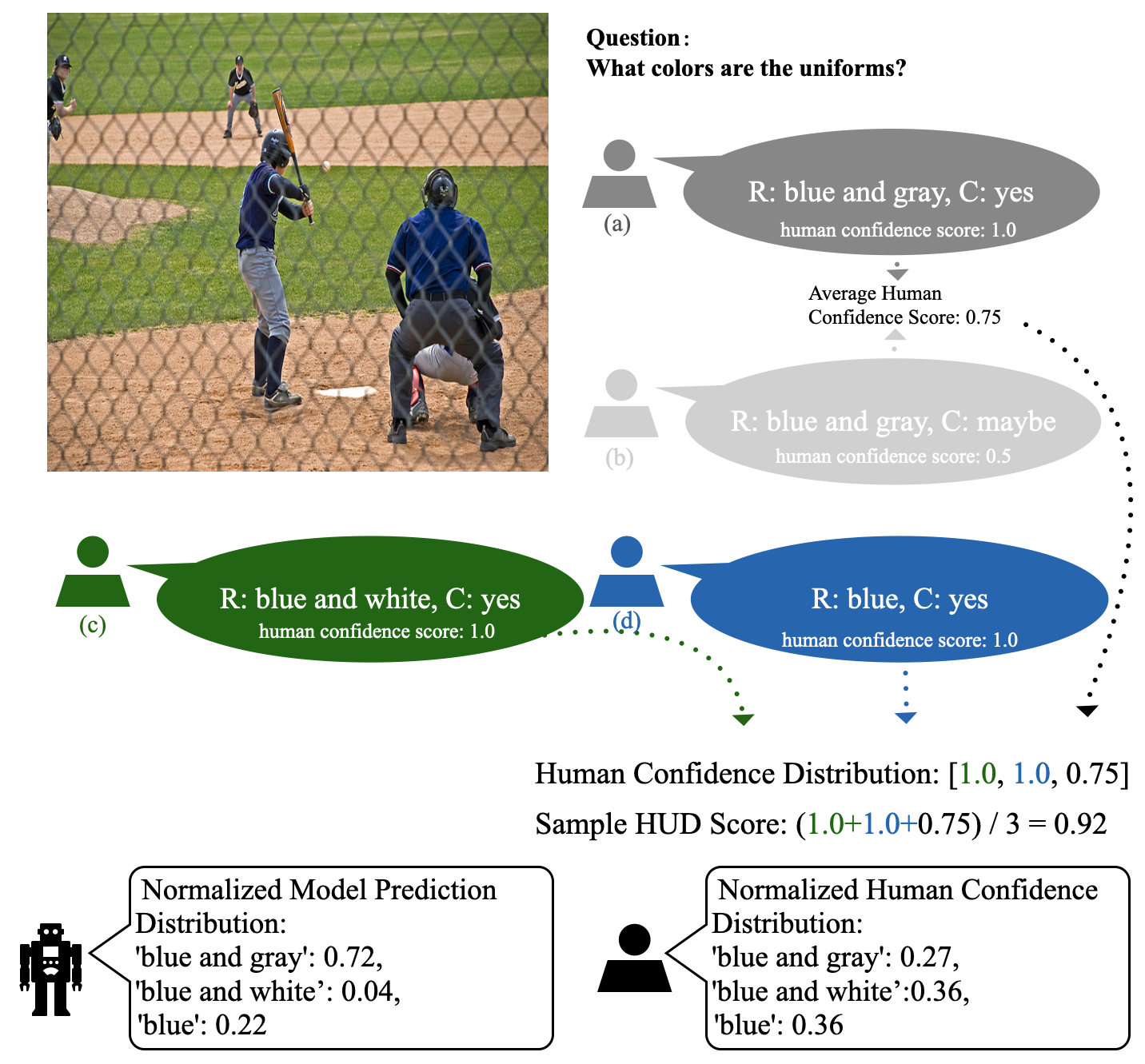}
    \caption{An example of VQA 2.0. We show different people's annotation in different colors, and compare the model predictions with human distributions. Please note there are 10 annotators for each sample in VQA 2.0, we only use four in the figure as a clear and easy display, where R is human response, and C is the confidence label.}
    \label{fig1}
\end{figure}

Large vision-language models should exhibit reliable model confidence so that humans can trust their outputs. Regardless of high model accuracy, poor confidence would indicate that a model does not know when it does not know \cite{stop, pavlick2019inherent, yang2024maqa}. Addressing the importance of model confidence has drawn increasing attention in recent years \cite{naeini2015obtaining, guo2017calibration}, aiming to make models more reliable and capable of mirroring humans' uncertainty \cite{uncertainty_acm, humans-exhibit-uncertainty,unc_rob}, especially for cases where human disagreement exists. We here refer to human uncertainty in human disagreement (HUD) as a two-part observation: a) the fact that for each sample, different individuals might have different knowledge, and thus it may lead to different (valid) responses (also known as \textit{human label variation} 
\cite{plank-2022-problem}); we here complement that with a second aspect, namely b) that the same humans may also have their own degrees of uncertainty in responding. HUD is a huge challenge that makes it difficult for models to align with humans \cite{unc_rob}, as many samples may not have a single ground truth with complete human agreement and high confidence. This issue is particularly pronounced in the visual question answering (VQA) task, where humans are observed to frequently have different opinions, and are easy to be uncertain with their answers. 
Figure \ref{fig1} shows an example of the task and the responses generated by the human annotators (R). It also provides the confidence labels (C) which express human uncertainty, where each annotator indicates their level of confidence in the given response. In the example, the two responses in (a) and (b) are the same (``blue and gray'') but with different human confidence labels (1.0 vs.\ 0.5), while those in (c) and (d) have different response labels with the same human confidence labels. Traditionally, a model would predict the response with the highest model prediction probability (e.g., ``blue and gray") but it would ignore the effect that human confidence plays on the prediction distribution. It will also miss the existence of multiple valid responses to the same question. For example, the answer ``blue" is less probable in the model prediction, but in reality it has a high human confidence and is indeed a highly possible answer. Therefore, training a model to predict the most probable correct answer, while ignoring HUD, does not give models the ability to accurately reflect human behavior and distributions in real-world scenarios. It also makes models struggle to assign reliable probabilities to other valid answers (``blue'' being 0.22). Most of the previous VQA studies have not explicitly addressed HUD, and have predominantly evaluated models based on accuracy of a single prediction of each sample alone.

This study examines how well state-of-the-art VQA models align with human confidence distributions when HUD exists. Unlike previous work on VQA, we propose to explicitly utilize HUD information to evaluate the discrepancies between model predictions and human responses. Specifically, we use VQA~2.0 \cite{vqa2.0}, a dataset where ten annotators answered each question and also indicated their confidence level in their answer. In this study we aim to answer the following research questions: $\textbf{RQ1:}$ To what extent do different levels of HUD impact the accuracy of the model and its alignment with human confidence? $\textbf{RQ2:}$ Does  calibration improve the human-model alignment? 

To answer the research questions, we start by adopting a splitting strategy, categorizing samples into three sets (low, medium, high) based on their levels of HUD. We then employ not only traditional accuracy but also propose to use three human-correlated metrics for VQA: Total Variation Distance (\textit{TVD}), Kullback-Leibler divergence (\textit{KL}), and Human Entropy Calibration Error (\textit{EntCE}), to comprehensively analyze the discrepancies between models and human in the context of HUD. Furthermore, we apply Temperature Scaling \cite{guo2017calibration} to verify the effectiveness of the traditional calibration method. 

Our results show that even state-of-the-art models such as BEiT3 \cite{beit} struggle to capture the multi-label distribution due to diverse human responses. Additionally, we observe that commonly used accuracy-oriented calibration techniques adversely affect BEiT3's confidence, further widening the gap between model prediction distributions and human confidence distributions. Instead, we propose to calibrate models towards human distributions, and thus better align models with humans under HUD. Our key contributions are: \textbf{1)} To the best of our knowledge, we are the first to study human uncertainty in disagreement in VQA, investigating the gap between the SOTA model prediction distributions and human response distributions; \textbf{2)} We are the first to implement a novel comprehensive evaluation of VQA models, evaluating on various human-centered metrics rather than solely relying on the VQA accuracy. We demonstrate that, when intrinsic HUD exists, the VQA accuracy can not adequately indicate a model's capacities; finally, \textbf{3)} We demonstrate that the variations in HUD levels affect model performances differently. We therefore advocate for an increased future attention on improving the alignment between models and humans under HUD.

\section{Related Work}

\paragraph{Human Disagreement:} Human disagreement, also known as human label variation \cite{plank-2022-problem}, has been studied in various natural language processing, computer vision, and human-computer interaction tasks, including natural language inference \cite{stop}, machine translation \cite{popovic-2021-agree}, question answering \cite{kwiatkowski-etal-2019-natural}, image classification \cite{unc_rob}, and social computing \cite{disagreement2021}. All these studies are based on datasets where multiple human annotators give \textit{different plausible answers} for each sample \cite{nie}. They then either aggregate human labels, using majority vote for training, or using human soft labels \cite{unc_rob}, optimizing models towards a human distribution based on label frequency. Human uncertainty here means uncertainty reflected by annotators choosing different labels, and is different from annotators providing indicators of uncertainty in their response. Human uncertainty in responses so far is an understudied phenomenon in relation to human label variation.

\paragraph{Visual Question Answering:} The VQA task was first proposed by \citet{vqaacc}, where they provide a benchmark including the VQA 1.0 dataset and the tradition evaluation metric VQA-Accuracy (VQA-Acc). VQA 2.0 \cite{vqa2.0} is based on VQA 1.0, where commonsense questions are removed with the aim to force the model to uniquely rely on the image content for answering.

For this task, large pre-trained vision-language models have become mainstream \cite{lxm, mplug, vlmo, beit} and have been further improved by optimizing for VQA-Accuracy. Among them, LXMERT is a comparatively light model which provides the first baseline in the era of pre-trained models. It yields a VQA-Accuracy of 72.5 on VQA 2.0 test set. BEiT3 \cite{beit} is the latest and strongest SOTA model with a VQA-Accuracy of 84.03. In VQA, human disagreement is studied implicitly by \citet{ease}, where they propose a diagnose tool to identify a sample's difficulty level by calculating the numbers of different response labels among human annotators and clustering the semantics of responses based on Word2Vec \cite{mikolov2018advances}. However, to the best of our knowledge, human uncertainty in responses has not been yet studied in VQA, where most work only focuses on VQA-Accuracy as the sole evaluation metric.

\paragraph{Uncertainty in VQA:} VQA is a complex vision-language task and, for this reason, it is normal that a human annotator might be uncertain in providing a label. The level of uncertainty in the responses indicates that the annotators are not completely sure about a given label.
Unfortunately, only very few datasets explicitly collect and release annotation on human uncertainty. 
Two positive examples are VQA 2.0 \cite{vqa2.0} and VizWiz \cite{gurari2018vizwiz}, two widely used datasets for VQA. They contain both labels and uncertainty values generated by ten different human annotators (``How certain are you?'' ``yes'', ``maybe'', ``no'').

Despite the availability of this data, previous studies on VQA \cite{vqaacc,vqa2.0, gurari2018vizwiz, lxm, beit} have not directly addressed human uncertainty in the responses. They simply trained models towards the most frequent label, optimizing for the highest VQA-Accuracy. Only recently, model uncertainty towards multiple answers have become an emerging direction \cite{yang2024maqa}, where the community is focusing on measuring the level of uncertainty of large language models \cite{xiong2024can}. However, the main challenge is that, still, human uncertainty information is not often available, making it difficult to evaluate how models correlate with humans.

\section{Human Uncertainty in Disagreement}
We first introduce the VQA 2.0 samples used in our study, and then discuss how we calculate the HUD scores and group the samples in three HUD levels based on their scores. 

\subsection{Data}
In this study we use VQA 2.0 \cite{vqa2.0}, one of the most commonly used VQA dataset.
Given that both BEiT3 \cite{beit} and LXMERT \cite{lxm} are pre-trained and fine-tuned only on VQA 2.0, this is the perfect dataset for a fair comparison between these two models. 

As shown in Figure \ref{fig1}, given an image-question-answer triplet $t=(i,q,\mathcal{A})$, where $i$ is an image, $q$ is a question, and $\mathcal{A}$ is an answer set, $\mathcal{A}$ consists of 10 independent humans' annotations. In each annotation $h_n$, $n = 1, \ldots, 10$, every annotator gives their response $r_n$ and also a confidence level $c_n$. The level $c_n$ corresponds to one of three pre-defined categories $<\text{`}yes\text{'}, \text{`}no\text{'}, \text{`}maybe\text{'}>$, indicating whether an annotator is confident in their answer.

\subsection{The HUD Score}
Since the confidence labels `yes', `no', and `maybe' are expressed in natural language, they can not be directly utilized for comparisons with model prediction distributions. 
To quantify human uncertainty, we assign different human confidence scores to each response based on its confidence label, as shown in Figure \ref{fig1}. We quantify every `yes' as 1.0, `no' as 0.01, and `maybe' as 0.5 respectively. We want the values of the three categories to be restricted between 0 and 1, as this range aligns with the distribution of the model's predicted probabilities. Additionally, in this work we assume that the value of `maybe' is the average score of `no' and `yes', ensuring it does not bias towards either end. We assign a very small value to `no' to avoid the situation of a distribution of all zeros for which entropy cannot be calculated. 

For the same response label with multiple annotations, for example, in Figure \ref{fig1} there are two people both answering `blue and gray', we calculate their average confidence scores as the human confidence score for this answer, and thus we get the human uncertainty scores for each response label. We finally calculate the average across all human response labels as the HUD score for this sample.

\subsection{Grouping samples based on HUD Levels}
\label{rank}

Our hypothesis is that a model would perform differently according to the different HUD level of the samples. For this reason, we divide the samples into three equal portions according to their HUD scores, namely low human uncertainty set (we use `low' set in short for the rest of the paper), medium human uncertainty set (medium set), and high human uncertainty set (high set), where humans are more certain about the samples in low set, while more uncertain in the high set.

\section{Model Confidence Scores and Calibration}
Here we discuss how we obtain the model prediction distributions and how to better align them with humans.

\subsection{Model Prediction Distribution}
Given a sample $x$, the prediction distribution of a model $\mathcal{M}$, denoted as $\mathcal{M}(x)$, corresponds to the probabilities $\text{P}_Y(Y=y|X=x)$ the model assigns to each class $y$. The most standard and most commonly used method to gain $\text{P}_{Y}$ is to extract the last hidden state of the neural network for each label, and use the $softmax$ function, which converts a vector of raw scores (logits), into a probability distribution over multiple classes to get the final normalized probability distribution: $\text{P}_{Y} = softmax \text{([}l_1, l_2, ..., l_K\text{])}$, where $K$ is the number of all labels.

\subsection{Calibration Method}
Calibration aims to adjust the predicted probabilities $\text{P}_Y$ towards a more reliable distribution \cite{guo2017calibration, desai-durrett-2020-calibration, jiang-etal-2021-know}. A well calibrated multi-class model accurately captures the true likelihood of predictions for all possible classes \cite{stop, vaicenavicius2019evaluating, kull2019beyond}. For instance, in VQA, if there are four possibly correct answers for a question and the humans' uncertainty scores for them are: [0.60, 0.30, 0.05, 0.05] respectively, a well calibrated model $\mathcal{M}$ should also have a prediction distribution approximately close to [0.60, 0.30, 0.05, 0.05] for same four labels. 

Among different calibration techniques, Temperature Scaling \cite{guo2017calibration} is one of the most widely used and most traditional methods. Temperature Scaling is a post-processing technique applied to the logits of a network by dividing them by a temperature parameter $T > 0$. Then we can use the modified logits and the $softmax$ function to get a calibrated distribution: $P_{Y=y} = \frac{e^{(l_y / T)}}{\sum_{j=1}^{K}e^{l_j / T}}$. In this work, we employ Temperature Scaling to test if this traditional calibration method is effective in our experiments, whether it is more effective for either of the two models, and indicate what a good calibration strategy is.

\section{Experiments}
\paragraph{Models:}
Our study focuses on the latest SOTA model BEiT3 \cite{beit} and the previous SOTA, also one of the most commonly used models LXMERT \cite{lxm}. We target on evaluating their differences on different HUD sets.

\paragraph{Fine-Tuning Set and Implementation Details:} The open-sourced training set for VQA 2.0 includes 443,757 samples, and the validation set has 213,954 samples. The test set is not open-sourced and the human uncertainty labels are not available. Therefore, we exclude the test set from our experiments. We follow exactly the same implementation details provided by BEiT3\footnote{https://github.com/microsoft/unilm/tree/master/beit3} and LXMERT\footnote{https://github.com/airsplay/LXMERT}, using their checkpoints (for BEiT3, we use the BEiT3-base model) of the pre-trained models and fine-tune them following the original instructions. More implementation details are provided in the Appendix A. It is essential to point out that BEiT3 and LXMERT utilize slightly different sets for fine-tuning and validation. They partitioned the original validation set in a customized manner, where BEiT3 keeps 5,303 samples from the original validation set and add the remaining set together with the training set for fine-tuning, while LXMERT keeps 25,994 samples from it and do the same. Even though this setting is a bit different, in previous works they do not strictly restrict this when reporting and comparing model performances. Therefore, we also keep the original setting as initially proposed.

\paragraph{Validation Set Details:} \label{val} 
We use BEiT3's 5,303 samples and LXMERT's 25,994 samples for validation, since neither of models have been fine-tuned on these samples before. We also filter out all the samples with only one human response label (no disagreement). These samples do not have a human confidence distribution, and thus their entropy can not be computed.

We are left with a set of 3,248 samples for BEiT3, where there are 1,083, 1,083, 1,082 samples for the low, medium, and high set respectively. Similarly, we have a set of 15,408 samples for LXMERT, where there are 5,136 samples for each HUD set.

\begin{figure}[ht!]
    \centering
    \includegraphics[width=\linewidth]{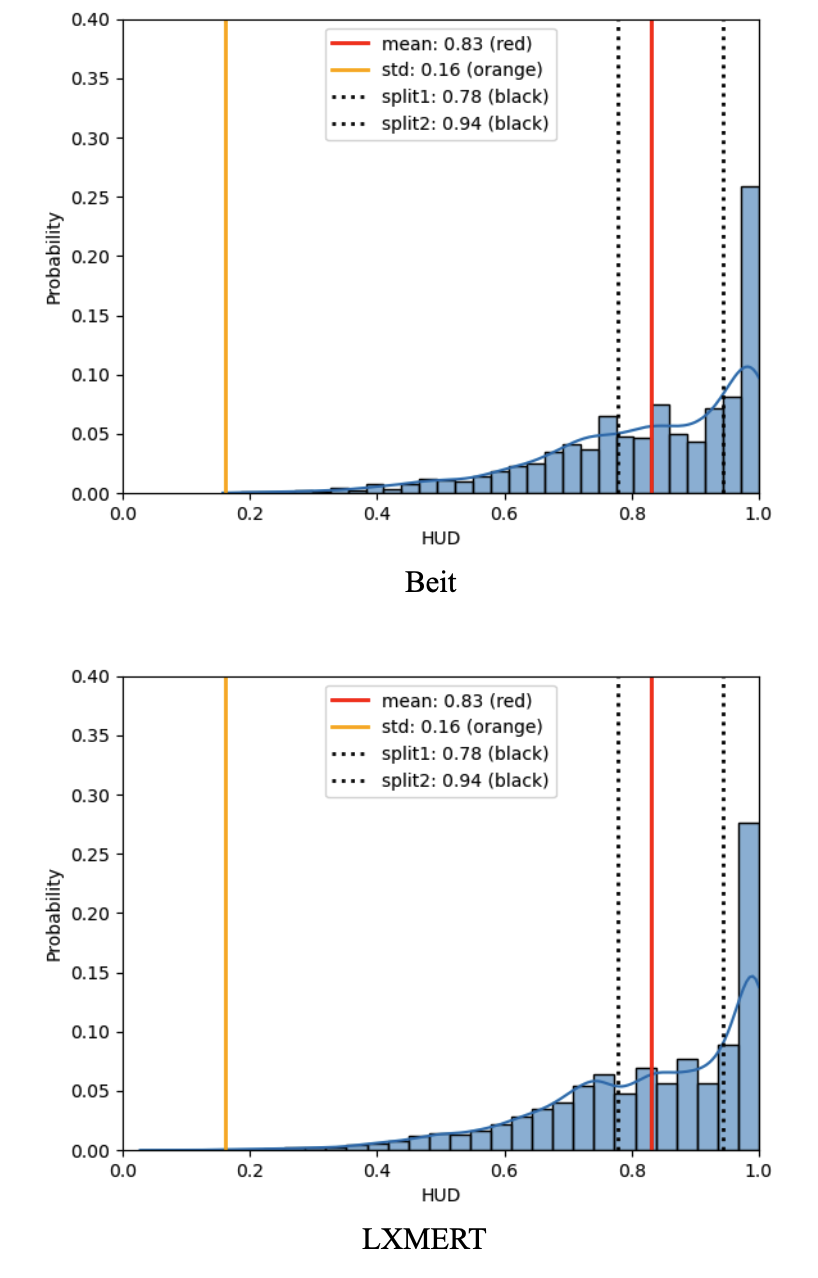}
    \caption{The data distributions of the two validation sets based on HUD scores. We show the two split boundaries using the black lines, the standard variation value in yellow lines, and mean values in red lines.}
    \label{fig2}
\end{figure}

For BEiT3, the average number of distinct human responses per sample in low set is 3.06, with an average HUD score of 0.98. In the medium set, these values are 3.97 and 0.86, respectively. While in the high set, they are 4.82 and 0.64, respectively. Similarly, for LXMERT, for low, medium, and high set, the values are: 3.08 and 0.98, 4.04 and 0.86, 4.67 and 0.64, respectively. Figure~\ref{fig2} shows two very similar distributions of HUD scores for the two validation sets, where they have the same mean value and standard variation value (std). The two black lines in both figures show the split boundaries of the low, medium, and high sets. We point out that even though our HUD set split strategy causes seemingly high confidence scores on each set, we target on revealing the differences in model performances caused by three sets, where humans have comparatively high, medium, and low uncertainty. Also, since the data distributions are similar, the two models are still comparable, even though our main focus is not to compare the two models' performances against each other. 

\paragraph{Evaluation metrics:} 
To conduct our evaluation, we include the following metrics. The specific equations are in Appendix A. 

\par
\textbf{VQA-Accuracy}: 
Given a model prediction, VQA-Accuracy \cite{vqaacc} is defined as:

\begin{equation}
\text{Acc}(\text{ans}) = \min \left\{ \frac{\# \text{humans that said } \text{ans}}{3}, 1 \right\}.
\label{vqaacc}
\end{equation}

VQA-Accuracy takes into account human label frequency, assigning higher scores to answers annotated by a greater number of human annotators. It is maximized (1.0), if at least 3 raters gave the exact answer. The number 3 is a manually set parameter proposed in the original work \cite{vqaacc}.

However, it does not consider human uncertainty scores. In this evaluation scenario, a model response overlapping with an annotator response with a very low confidence would still be considered correct.

\textbf{KL-divergence (KL)} \cite{kullback1951information} measures how the model probability distribution diverges from human's, denoted as KL(Human$||$Model). Given a human distribution, the model distribution is the approximate distribution, and the KL score indicates how much the approximation distribution deviates from the true distribution. \par

\textbf{Total Variation Distance (TVD)} \cite{devroye2001combinatorial} measures the absolute difference between the model's and human's probability distributions, to indicate how much overlap there is between two distributions. In \citet{stop}, this measure was proposed as Human Distribution Calibration Error.

\textbf{Human Entropy Calibration Error (EntCE)} measures the absolute difference between human's and a model’s entropy of the their distribution, and indicate the average confusion difference in two distributions \cite{stop}.

\par
\subsection{Results and Discussion}

\begin{table}[tbp]
    \centering
    \resizebox{\columnwidth}{!}{
    \begin{tabular}{ll|c|ccc}
        \toprule
        Model & \makecell[l]{HUD \\ set split} & VQA-Acc $\uparrow$ & TVD $\downarrow$ & KL(H\textbar\textbar M) $\downarrow$ & EntCE $\downarrow$ \\
        \midrule
        \multirow{4}{*}{LXMERT} 

        & All   & 0.671              & 0.448             & 1.886             & 0.555 \\
        & Low   & \textbf{0.761}     & 0.454             & 2.237             & 0.580 \\
        & Med   & 0.675              & 0.450             & 1.786             & 0.579 \\
        & High  & 0.576              & \textbf{0.439}    & \textbf{1.634}    & \textbf{0.506} \\
                               
        \midrule
        \multirow{4}{*}{BEiT3}  
        & All   & 0.842              & 0.436             & 1.701             & 0.541 \\
        & Low   & \textbf{0.921}     & 0.455             & 2.150             & 0.575 \\
        & Med   & 0.845              & 0.437             & 1.544             & 0.546 \\
        & High  & 0.759              & \textbf{0.416}    & \textbf{1.408}    & \textbf{0.503} \\
                               
        \bottomrule
    \end{tabular}
    }
    \caption{Model performances on VQA-Accuracy and three Human-correlated metrics before calibration. For each model, the \textbf{best results} are highlighted in bold. HUD: High means human express high uncertainty for their responses.}
    \label{tab1}
\end{table}

\begin{table*}[tbp]
    \centering
        \begin{tabular}{lll|llll|lll}
\midrule
& & & \multicolumn{4}{c|}{\makecell{Calibration towards \\ VQA-Acc}} & \multicolumn{3}{c}{\makecell{Calibration towards \\ Human Distributions}} \\  

\multirow{2}{*}{Model} & \makecell[l]{HUD \\ set split}& \makecell[l]{ECE \\ Before} $\downarrow$  & \makecell{ECE \\ After} $\downarrow$ & TVD $\downarrow$ & KL $\downarrow$ & EntCE $\downarrow$  & TVD $\downarrow$ & KL $\downarrow$ & EntCE $\downarrow$ \\ \midrule

\multirow{4}{*}{LXMERT} 
    & All  & 0.066 & 0.065 & 0.442 & 1.771 & 0.542 & 0.354 & 0.731 & 0.353 \\
    & Low  & 0.072 & 0.070 & 0.449 & 2.103 & 0.569 & 0.369 & 0.877 & 0.397 \\
    & Med  & 0.069 & 0.069 & 0.444 & 1.674 & 0.564 & 0.347 & 0.667 & 0.351 \\
    & High & 0.071 & 0.066 & 0.433 & 1.534 & 0.492 & 0.345 & 0.649 & 0.310 \\ \midrule
\multirow{4}{*}{BEiT3} 
    & All  & 0.094 & 0.039 & 0.488 (-) & 3.232 (-) & 0.670 (-) & 0.340 & 0.647 & 0.333 \\
    & Low  & 0.084 & 0.039 & 0.496 (-) & 3.999 (-) & 0.674 (-) & 0.371 & 0.837 & 0.389 \\
    & Med  & 0.099 & 0.042 & 0.494 (-) & 2.982 (-) & 0.683 (-) & 0.331 & 0.563 & 0.318 \\
    & High & 0.098 & 0.046 & 0.473 (-) & 2.712 (-) & 0.655 (-) & 0.319 & 0.542 & 0.292 \\ \hline
\end{tabular}
    \caption{Model performances on ECE, TVD, KL(Human $||$ Model), and EntCE after calibration. We compare the calibration direction towards accuracy and towards human. We use (-) to indicate when calibration negatively affects the model performance.}
    \label{tab2}
\end{table*}

\textbf{Before Calibration.} Table~\ref{tab1} reports the performances of the two models (without calibration) across three different levels of human uncertainty, as well as for the overall evaluation set. Performance metrics include VQA-Accuracy, TVD, KL, and EntCE. Overall, we observe that BEiT3 consistently outperforms LXMERT in accuracy (0.67 vs. 0.84), thereby confirming that it is the better model in terms of accuracy. When zooming into the three HUD levels, both models show a similar trend: as human uncertainty level increases from low to high, their accuracy decreases. The models obtain the highest accuracy level for the stimuli in the low HUD set, as humans provide answers with high confidence to the stimuli in that category. The continuous decline in both models' VQA-Accuracy performance supports our hypothesis that the higher the HUD level is (more uncertainty), the more challenging it is for a model to predict the best answer.

Besides accuracy, we look at model confidence and compare it against human label variation (which is now across raters). We provide models' performances on TVD, KL, and EntCE to reveal the extent to which the model prediction distributions approximate humans'. When we compare the two models, both on the overall set and on the three subsets, the results on all three metrics show that BEiT3 correlates better with humans than LXMERT. For both models, we surprisingly find they achieve their best performances on high set (high human uncertainty). We provide an in-depth analysis as follows. 

As introduced in the validation set details, on average for both models, there are around 3.0, 4.0, and 4.8 different human response labels for a sample in low, medium, and high set, with 0.98, 0.86, 0.64 HUD scores respectively. This indicates that the more uncertain humans are, the higher is the human disagreement. We believe the reason both models have the highest scores on the low set is that the samples have comparatively fewer human response labels, and thus it is easier for a model to learn a `seemingly most correct answer' but ignore the other possibilities. This is also why models are having higher accuracy on the low set; it learns to predict the most correct answer which is supported by the human majority vote, but it does not learn to align well with the overall human answer distribution, i.e., on all the other possible answers. On the contrary, for samples from the high human uncertainty set, there are comparatively more response labels with a smoother confidence distribution (less abrupt variations between the confidence scores on each label), where it helps the models to pay attention to possible labels rather than focusing on one single label. We further showcase and analyze this in the Case Study.

\paragraph{Calibration towards VQA-Accuracy} Table~\ref{tab2} shows the effectiveness of Temperature Scaling. The effect of TS is traditionally evaluated by \textbf{Expected Calibration Error (ECE)} \cite{guo2017calibration}, which measures the absolute difference between the accuracy of the predictions of a model and its confidence towards these predictions. This evaluation quantifies how well the model's predicted probabilities are calibrated towards its accuracy. In Table~\ref{tab2} we report the ECE results before and after TS. We empirically set the Temperature $t$ by comparing the $t$ from 0.1 to 2.0 with an interval of 0.05. We set $t$ to 1.05 for LXMERT and 0.6 for BEiT3, where they each have the lowest ECE on the overall set after TS. We then fix $t$ and report the results for TVD, KL, and EntCE. As expected, the ECE scores of all sets drop after calibration. However, using TS towards VQA-Accuracy undermines BEiT3's confidence (indicated by the minus symbol`(-)'), where TVD, KL, and EntCE scores become higher e.g., from 1.701 to 3.232 on KL for BEiT3. On the contrary, calibrating towards VQA-Accuracy only helps LXMERT improve the correlation with the three human distributions. We conclude that the traditional calibration technique still works for slightly weaker models like LXMERT, but they do not help the latest strong model BEiT3. Besides, here we do not further compare the ECE results between different sets or between models, but only use ECE to test if using TS to calibrate towards VQA-Accuracy works and how strongly this technique influences the models' correlation with human. ECE does not consider human disagreement and distributions, and thus we do not further analyze ECE results.

\paragraph{Calibration towards human distributions.} 
A crucial aspect of this work involves using TS as a calibration method to align with human distributions by reducing scores on TVD, KL, and EntCE. Similar to what we introduced in the previous paragraph, we empirically set the temperature $t$ to be 2.0 for both models, as it helps model reach better results. Normally, $t$ is not supposed to be a large value since a too high value makes the predicted probabilities become closer to a uniform distribution, reducing the differences between the classes, which is not desired. As shown in Table \ref{tab2} on the right-hand side, on the overall set and also three subsets, calibrating towards human distributions rather than VQA-Accuracy helps both models reach lower scores (better results) on TVD (e.g. from 0.448 to 0.354 on LXMERT, from 0.436 to 0.340 on BEiT3), KL (e.g. from 1.886 to 0.731 on LXMERT, from 1.701 to 0.647 on BEiT3), and EntCE (e.g. from 0.555 to 0.353 on EntCE, from 0.541 to 0.333 on BEiT3) compared with the original results in Table \ref{tab1}. Notably, TVD and EntCE measure the average discrepancies between two distributions (on each element in the vector), but are not sensitive to element class rankings. KL is sensitive to class rankings, where small changes in probabilities can lead to large increases in the divergence. Therefore, we observe a drastic decrease in KL.

On each set split, BEiT3 still outperforms LXMERT on all three metrics, while both models reach the best performances on the high set, and the worst performances on low set. We conclude that, even though BEiT3 has a seemingly strong performance on all metrics, it is not yet correlated well with humans, especially on the low HUD set. Moreover, by calibrating towards human distributions, it even yields very similar human correlation performances when compared with LXMERT. This indicates that BEiT3 is simply better optimized towards human majority vote based on label frequency, but it is not much better at learning true human distribution. In other words, while BEiT3 is the better model in terms of accuracy compared to LXMERT, BEiT3 is less well-aligned to human preferences. This opens up for interesting future research directions on how to strike a good balance between training a high-accuracy model and one that is well aligned with humans.

\subsection{Case Study}

\begin{figure*}[ht!]
    \centering
    \includegraphics[width=\linewidth]{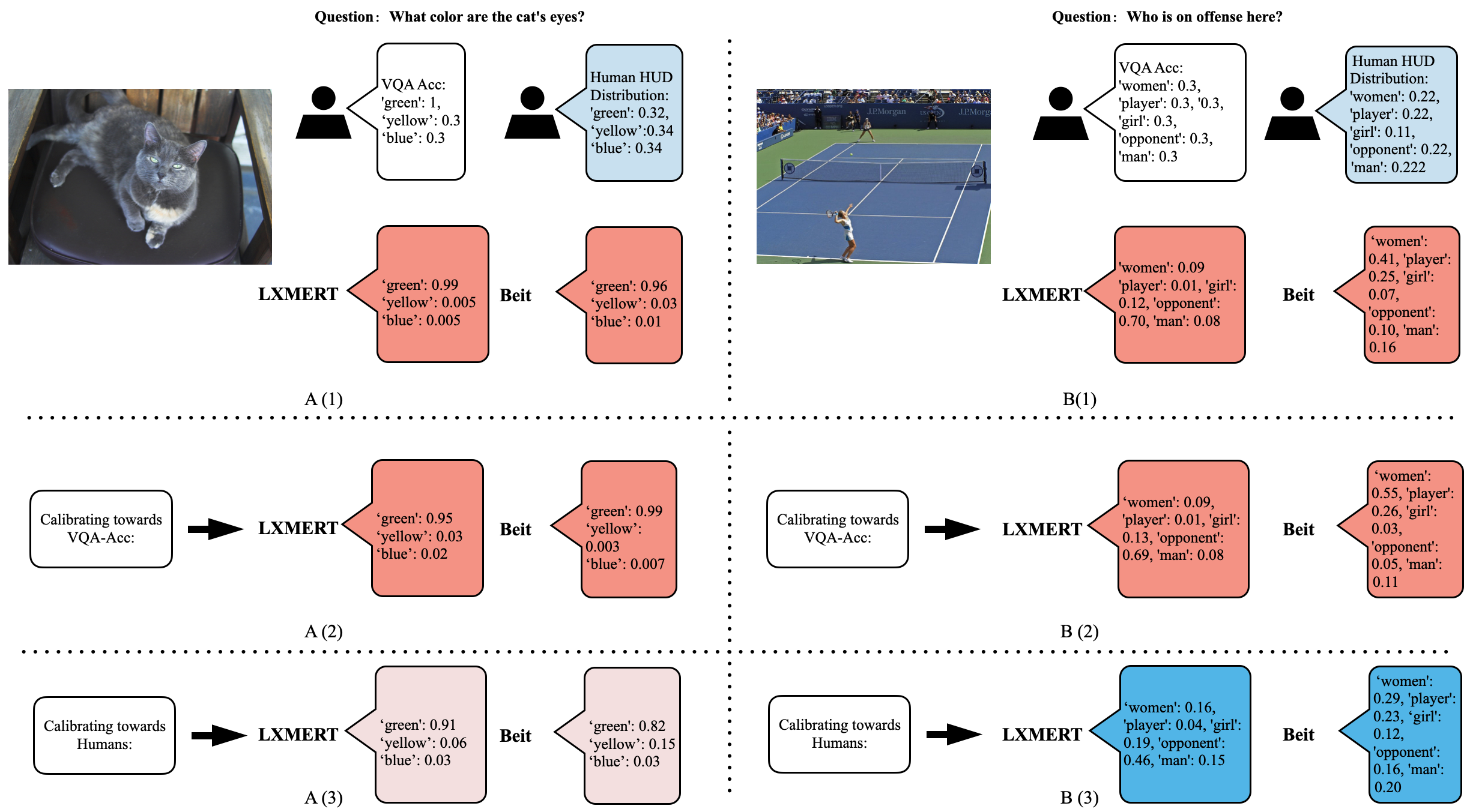}
    \caption{Case Study. 
    The left part (A(1)-A(3)) presents a sample from the \textbf{low set}, displaying each response label’s VQA Accuracy score, Human HUD Distribution, and the prediction distributions of LXMERT and BEiT3.
    We report all three situations, where models are: not calibrated, calibrated towards VQA-Accuracy, and calibrated towards human distributions. Similarly, the right part (B(1)-B(3)) presents a sample from the \textbf{high set}. Human HUD distributions, which are used for measuring human-model correlation, are colored in light blue. In red we highlight unsatisfactory model performances, in light red models that slightly improved but are still not good, while dark blue indicates much better correlation with humans.}
    \label{fig3}
\end{figure*}

In Figure~\ref{fig3} we showcase two samples from the low (A(1)-A(3)) and high (B(1)-(B3)) sets and compare models' specific predictions against humans'. In line with what we discussed in the last section, on the low set (see A(1)), the model approximates only towards the most correct answer `green', which has the highest VQA-Accuracy. However, besides `green', in the human distribution, `yellow' and `blue' also have similar weights. This means the model should predict all three answers with similar distributions to mirror humans' confidence in the real world. However, both BEiT3 and LXMERT show an extremely high probability for `green', and very low probabilities for `yellow' and `blue'. When calibrating towards VQA-Accuracy, in A(2) we see a very slight improvement in LXMERT. Once again, this traditional calibration method negatively affects the performance of BEiT3.
On the contrary, when calibrating towards humans (see A(3)), there is a continuous improvement in both models, even though they are still far from human performance.
On the right side (see B(1)), the sample from the high set contains more human response labels, each with lower but similar VQA-Accuracy scores.
In this case, BEiT3 distributes the probability mass across most of the labels provided without the huge gap seen in A(1) (0.96 vs 0.01).
Although LXMERT also assigns probability scores to multiple labels, it strongly favors one label ('opponent') with a probability much higher than the others.
In B(2), calibrating towards VQA-Accuracy causes tiny changes on the labels `girl' and `opponent', while it again does harm BEiT3, causing much higher values for `women', while decreases the probabilities for `opponent' and `man'. On the contrary, calibrating towards humans work well for both models, making them both correlate better with humans. 

Finally, by comparing the results in A(3) and in B(3) we see how even when calibrating towards humans, the models still predict one specific label (e.g., `green') while mostly ignoring the other candidates (e.g., `yellow' and `blue'). This clearly show why models show a better correlation with humans on the high set.

\section{Conclusion}
In this work, we study human uncertainty in Visual Question Answering. We find that when humans exhibit high levels of uncertainty in disagreement, the models have difficulty in predicting the most correct answer, but are actually better at correlating with human confidence distributions. We further find that the traditional calibration strategy towards accuracy does not work for BEiT3, the latest SOTA model. Instead, we then demonstrate that calibrating it towards human distribution is more effective. We conclude that evaluating VQA models uniquely on the VQA-Accuracy metric is not sufficient, and future studies should focus more on aligning models with human uncertainty and disagreement.    

\section{Limitation}
Our study has three limitations. Firstly, while being highly reproducible, our study is restricted to two slightly different validation sets, even though we believe they are much comparable. Future study could extend our study to fine-tuning and validating models on more datasets. Secondly, it is not feasible to explore all the strategies to quantify human uncertainty labels. Although our study provides a very standard solution, this represents 
a good start point; we advocate for future research to explore more advanced quantification strategies. Thirdly, in our work we use Temperature Scaling; future studies should consider other calibration techniques to directly target human uncertainty in disagreement.

\section{Ethics Statement}
We anticipate no ethical concerns with this work. We utilized open-sourced data and models, which have been appropriately cited.

\nobibliography*

\bibliography{aaai25}

\begin{thebibliography}{29}
\providecommand{\natexlab}[1]{#1}

\bibitem[{Antol et~al.(2015)Antol, Agrawal, Lu, Mitchell, Batra, Zitnick, and Parikh}]{vqaacc}
Antol, S.; Agrawal, A.; Lu, J.; Mitchell, M.; Batra, D.; Zitnick, C.~L.; and Parikh, D. 2015.
\newblock Vqa: Visual question answering.
\newblock In \emph{Proceedings of the IEEE international conference on computer vision}, 2425--2433.

\bibitem[{Baan et~al.(2022)Baan, Aziz, Plank, and Fernandez}]{stop}
Baan, J.; Aziz, W.; Plank, B.; and Fernandez, R. 2022.
\newblock Stop Measuring Calibration When Humans Disagree.
\newblock In Goldberg, Y.; Kozareva, Z.; and Zhang, Y., eds., \emph{Proceedings of the 2022 Conference on Empirical Methods in Natural Language Processing}, 1892--1915. Abu Dhabi, United Arab Emirates: Association for Computational Linguistics.

\bibitem[{Bao et~al.(2022)Bao, Wang, Dong, Liu, Mohammed, Aggarwal, Som, Piao, and Wei}]{vlmo}
Bao, H.; Wang, W.; Dong, L.; Liu, Q.; Mohammed, O.~K.; Aggarwal, K.; Som, S.; Piao, S.; and Wei, F. 2022.
\newblock Vlmo: Unified vision-language pre-training with mixture-of-modality-experts.
\newblock \emph{Advances in Neural Information Processing Systems}, 35: 32897--32912.

\bibitem[{Collins et~al.(2023)Collins, Barker, Espinosa~Zarlenga, Raman, Bhatt, Jamnik, Sucholutsky, Weller, and Dvijotham}]{uncertainty_acm}
Collins, K.~M.; Barker, M.; Espinosa~Zarlenga, M.; Raman, N.; Bhatt, U.; Jamnik, M.; Sucholutsky, I.; Weller, A.; and Dvijotham, K. 2023.
\newblock Human Uncertainty in Concept-Based AI Systems.
\newblock In \emph{Proceedings of the 2023 AAAI/ACM Conference on AI, Ethics, and Society}, AIES '23, 869–889. New York, NY, USA: Association for Computing Machinery.
\newblock ISBN 9798400702310.

\bibitem[{Desai and Durrett(2020)}]{desai-durrett-2020-calibration}
Desai, S.; and Durrett, G. 2020.
\newblock Calibration of Pre-trained Transformers.
\newblock In Webber, B.; Cohn, T.; He, Y.; and Liu, Y., eds., \emph{Proceedings of the 2020 Conference on Empirical Methods in Natural Language Processing (EMNLP)}, 295--302. Online: Association for Computational Linguistics.

\bibitem[{Devroye and Lugosi(2001)}]{devroye2001combinatorial}
Devroye, L.; and Lugosi, G. 2001.
\newblock \emph{Combinatorial methods in density estimation}.
\newblock Springer Science \& Business Media.

\bibitem[{Gordon et~al.(2021)Gordon, Zhou, Patel, Hashimoto, and Bernstein}]{disagreement2021}
Gordon, M.~L.; Zhou, K.; Patel, K.; Hashimoto, T.; and Bernstein, M.~S. 2021.
\newblock The disagreement deconvolution: Bringing machine learning performance metrics in line with reality.
\newblock In \emph{Proceedings of the 2021 CHI Conference on Human Factors in Computing Systems}, 1--14.

\bibitem[{Goyal et~al.(2017)Goyal, Khot, Summers-Stay, Batra, and Parikh}]{vqa2.0}
Goyal, Y.; Khot, T.; Summers-Stay, D.; Batra, D.; and Parikh, D. 2017.
\newblock Making the v in vqa matter: Elevating the role of image understanding in visual question answering.
\newblock In \emph{Proceedings of the IEEE conference on computer vision and pattern recognition}, 6904--6913.

\bibitem[{Guo et~al.(2017)Guo, Pleiss, Sun, and Weinberger}]{guo2017calibration}
Guo, C.; Pleiss, G.; Sun, Y.; and Weinberger, K.~Q. 2017.
\newblock On calibration of modern neural networks.
\newblock In \emph{International conference on machine learning}, 1321--1330. PMLR.

\bibitem[{Gurari et~al.(2018)Gurari, Li, Stangl, Guo, Lin, Grauman, Luo, and Bigham}]{gurari2018vizwiz}
Gurari, D.; Li, Q.; Stangl, A.~J.; Guo, A.; Lin, C.; Grauman, K.; Luo, J.; and Bigham, J.~P. 2018.
\newblock Vizwiz grand challenge: Answering visual questions from blind people.
\newblock In \emph{Proceedings of the IEEE conference on computer vision and pattern recognition}, 3608--3617.

\bibitem[{Ilia and Aziz(2024)}]{humans-exhibit-uncertainty}
Ilia, E.; and Aziz, W. 2024.
\newblock Predict the Next Word: {\textless}Humans exhibit uncertainty in this task and language models {\_}{\_}{\_}{\_}{\_}{\textgreater}.
\newblock In Graham, Y.; and Purver, M., eds., \emph{Proceedings of the 18th Conference of the European Chapter of the Association for Computational Linguistics (Volume 2: Short Papers)}, 234--255. St. Julian{'}s, Malta: Association for Computational Linguistics.

\bibitem[{Jiang et~al.(2021)Jiang, Araki, Ding, and Neubig}]{jiang-etal-2021-know}
Jiang, Z.; Araki, J.; Ding, H.; and Neubig, G. 2021.
\newblock How Can We Know When Language Models Know? On the Calibration of Language Models for Question Answering.
\newblock \emph{Transactions of the Association for Computational Linguistics}, 9: 962--977.

\bibitem[{Jolly, Pezzelle, and Nabi(2021)}]{ease}
Jolly, S.; Pezzelle, S.; and Nabi, M. 2021.
\newblock {E}a{S}e: A Diagnostic Tool for {VQA} based on Answer Diversity.
\newblock In Toutanova, K.; Rumshisky, A.; Zettlemoyer, L.; Hakkani-Tur, D.; Beltagy, I.; Bethard, S.; Cotterell, R.; Chakraborty, T.; and Zhou, Y., eds., \emph{Proceedings of the 2021 Conference of the North American Chapter of the Association for Computational Linguistics: Human Language Technologies}, 2407--2414. Online: Association for Computational Linguistics.

\bibitem[{Kull et~al.(2019)Kull, Perello~Nieto, K{\"a}ngsepp, Silva~Filho, Song, and Flach}]{kull2019beyond}
Kull, M.; Perello~Nieto, M.; K{\"a}ngsepp, M.; Silva~Filho, T.; Song, H.; and Flach, P. 2019.
\newblock Beyond temperature scaling: Obtaining well-calibrated multi-class probabilities with dirichlet calibration.
\newblock \emph{Advances in neural information processing systems}, 32.

\bibitem[{Kullback and Leibler(1951)}]{kullback1951information}
Kullback, S.; and Leibler, R.~A. 1951.
\newblock On information and sufficiency.
\newblock \emph{The annals of mathematical statistics}, 22(1): 79--86.

\bibitem[{Kwiatkowski et~al.(2019)Kwiatkowski, Palomaki, Redfield, Collins, Parikh, Alberti, Epstein, Polosukhin, Devlin, Lee, Toutanova, Jones, Kelcey, Chang, Dai, Uszkoreit, Le, and Petrov}]{kwiatkowski-etal-2019-natural}
Kwiatkowski, T.; Palomaki, J.; Redfield, O.; Collins, M.; Parikh, A.; Alberti, C.; Epstein, D.; Polosukhin, I.; Devlin, J.; Lee, K.; Toutanova, K.; Jones, L.; Kelcey, M.; Chang, M.-W.; Dai, A.~M.; Uszkoreit, J.; Le, Q.; and Petrov, S. 2019.
\newblock Natural Questions: A Benchmark for Question Answering Research.
\newblock \emph{Transactions of the Association for Computational Linguistics}, 7: 452--466.

\bibitem[{Li et~al.(2022)Li, Xu, Tian, Wang, Yan, Bi, Ye, Chen, Xu, Cao, Zhang, Huang, Huang, Zhou, and Si}]{mplug}
Li, C.; Xu, H.; Tian, J.; Wang, W.; Yan, M.; Bi, B.; Ye, J.; Chen, H.; Xu, G.; Cao, Z.; Zhang, J.; Huang, S.; Huang, F.; Zhou, J.; and Si, L. 2022.
\newblock m{PLUG}: Effective and Efficient Vision-Language Learning by Cross-modal Skip-connections.
\newblock In Goldberg, Y.; Kozareva, Z.; and Zhang, Y., eds., \emph{Proceedings of the 2022 Conference on Empirical Methods in Natural Language Processing}, 7241--7259. Abu Dhabi, United Arab Emirates: Association for Computational Linguistics.

\bibitem[{Mikolov et~al.(2018)Mikolov, Grave, Bojanowski, Puhrsch, and Joulin}]{mikolov2018advances}
Mikolov, T.; Grave, {\'E}.; Bojanowski, P.; Puhrsch, C.; and Joulin, A. 2018.
\newblock Advances in Pre-Training Distributed Word Representations.
\newblock In \emph{Proceedings of the Eleventh International Conference on Language Resources and Evaluation (LREC 2018)}.

\bibitem[{Naeini, Cooper, and Hauskrecht(2015)}]{naeini2015obtaining}
Naeini, M.~P.; Cooper, G.; and Hauskrecht, M. 2015.
\newblock Obtaining well calibrated probabilities using bayesian binning.
\newblock In \emph{Proceedings of the AAAI conference on artificial intelligence}.

\bibitem[{Nie, Zhou, and Bansal(2020)}]{nie}
Nie, Y.; Zhou, X.; and Bansal, M. 2020.
\newblock What Can We Learn from Collective Human Opinions on Natural Language Inference Data?
\newblock In Webber, B.; Cohn, T.; He, Y.; and Liu, Y., eds., \emph{Proceedings of the 2020 Conference on Empirical Methods in Natural Language Processing (EMNLP)}, 9131--9143. Online: Association for Computational Linguistics.

\bibitem[{Pavlick and Kwiatkowski(2019)}]{pavlick2019inherent}
Pavlick, E.; and Kwiatkowski, T. 2019.
\newblock Inherent disagreements in human textual inferences.
\newblock \emph{Transactions of the Association for Computational Linguistics}, 7: 677--694.

\bibitem[{Peterson et~al.(2019)Peterson, Battleday, Griffiths, and Russakovsky}]{unc_rob}
Peterson, J.~C.; Battleday, R.~M.; Griffiths, T.~L.; and Russakovsky, O. 2019.
\newblock Human uncertainty makes classification more robust.
\newblock In \emph{Proceedings of the IEEE/CVF international conference on computer vision}, 9617--9626.

\bibitem[{Plank(2022)}]{plank-2022-problem}
Plank, B. 2022.
\newblock The {``}Problem{''} of Human Label Variation: On Ground Truth in Data, Modeling and Evaluation.
\newblock In Goldberg, Y.; Kozareva, Z.; and Zhang, Y., eds., \emph{Proceedings of the 2022 Conference on Empirical Methods in Natural Language Processing}, 10671--10682. Abu Dhabi, United Arab Emirates: Association for Computational Linguistics.

\bibitem[{Popovi{\'c}(2021)}]{popovic-2021-agree}
Popovi{\'c}, M. 2021.
\newblock Agree to Disagree: Analysis of Inter-Annotator Disagreements in Human Evaluation of Machine Translation Output.
\newblock In Bisazza, A.; and Abend, O., eds., \emph{Proceedings of the 25th Conference on Computational Natural Language Learning}, 234--243. Online: Association for Computational Linguistics.

\bibitem[{Tan and Bansal(2019)}]{lxm}
Tan, H.; and Bansal, M. 2019.
\newblock {LXMERT}: Learning Cross-Modality Encoder Representations from Transformers.
\newblock In Inui, K.; Jiang, J.; Ng, V.; and Wan, X., eds., \emph{Proceedings of the 2019 Conference on Empirical Methods in Natural Language Processing and the 9th International Joint Conference on Natural Language Processing (EMNLP-IJCNLP)}, 5100--5111. Hong Kong, China: Association for Computational Linguistics.

\bibitem[{Vaicenavicius et~al.(2019)Vaicenavicius, Widmann, Andersson, Lindsten, Roll, and Sch{\"o}n}]{vaicenavicius2019evaluating}
Vaicenavicius, J.; Widmann, D.; Andersson, C.; Lindsten, F.; Roll, J.; and Sch{\"o}n, T. 2019.
\newblock Evaluating model calibration in classification.
\newblock In \emph{The 22nd international conference on artificial intelligence and statistics}, 3459--3467. PMLR.

\bibitem[{Wang et~al.(2023)Wang, Bao, Dong, Bjorck, Peng, Liu, Aggarwal, Mohammed, Singhal, Som et~al.}]{beit}
Wang, W.; Bao, H.; Dong, L.; Bjorck, J.; Peng, Z.; Liu, Q.; Aggarwal, K.; Mohammed, O.~K.; Singhal, S.; Som, S.; et~al. 2023.
\newblock Image as a foreign language: Beit pretraining for vision and vision-language tasks.
\newblock In \emph{Proceedings of the IEEE/CVF Conference on Computer Vision and Pattern Recognition}, 19175--19186.

\bibitem[{Xiong et~al.(2024)Xiong, Hu, Lu, LI, Fu, He, and Hooi}]{xiong2024can}
Xiong, M.; Hu, Z.; Lu, X.; LI, Y.; Fu, J.; He, J.; and Hooi, B. 2024.
\newblock Can {LLM}s Express Their Uncertainty? An Empirical Evaluation of Confidence Elicitation in {LLM}s.
\newblock In \emph{The Twelfth International Conference on Learning Representations}.

\bibitem[{Yang, Yoo, and Lee(2024)}]{yang2024maqa}
Yang, Y.; Yoo, H.; and Lee, H. 2024.
\newblock MAQA: Evaluating Uncertainty Quantification in LLMs Regarding Data Uncertainty.
\newblock arXiv:2408.06816.

\end{thebibliography}

\section{Appendix}
\subsection{A. Evaluation Metrics}\label{A}

Besides the description in the main text, we provide the detailed equation of metrics used in our work. 

\textbf{The Total Variation Distance} \cite{devroye2001combinatorial} between two probability distributions \( P \) and \( Q \) is defined as:
\begin{equation}
\text{TVD}(P, Q) = \frac{1}{2} \sum_{x \in \mathcal{X}} |P(x) - Q(x)|.
\end{equation}
The TVD is robust to tiny changes in probabilities, and reflects the overall absolute differences between two distributions.

\textbf{The Kullback-Leibler Divergence} \cite{kullback1951information} between two probability distributions \( P \) and \( Q \) is defined as:
\begin{equation}
D_{\text{KL}}(P \parallel Q) = \sum_{x} P(x) \log \frac{P(x)}{Q(x)}.
\end{equation}
Notably, the $D_\text{KL}(P \parallel Q)$ is not equal to $D_\text{KL}(Q \parallel P)$. KL divergence is sensitive to small changes in the distribution, where the former distribution P in $D_\text{KL}(P \parallel Q)$ is a given true distribution and the Q is an approximation.

\textbf{The Human Entropy Calibration Error} \cite{stop} between two distributions is defined as:
\begin{equation}
\text{EntCE}(x) = H(f(x)) - H(\bar{\pi}(x)),
\end{equation}
where $f(x)$ is the model prediction distribution, and $\bar{\pi}(x)$ is the human confidence distribution. $H(\cdot)$ is the entropy of a distribution. It measures the alignment between humans and a model’s `indecisiveness' \cite{stop}.

\textbf{The Expected Calibration Error (ECE)} \cite{guo2017calibration} is defined as:
\begin{equation}
    \text{ECE} = \sum_{m=1}^M \frac{\|\text{B}_m\|}{N}\|\text{VQA-Accuracy}(\text{B}_m) - \text{conf}(\text{B}_m)\|,
\end{equation}
where $M$ is the number of bins, $N$ is the total number of samples, and VQA-Accuracy($\cdot$) and conf($\cdot$) is model's accuracy and its probability on its prediction. The ECE divide the samples into different bins, and calculate the difference between model's accuracy and confidence according to these bins.  

\subsection{Reproducibility}
Our experiments and results are highly reproducible. All experiments are implemented on four NVIDIA A100-SXM4-80G GPUs and an AMD EPYC 7763 64-Core CPU. We report the training hyper-parameters in Table \ref{tab3}. BEiT3 and LXMERT have different settings to reach the best performances. Our selected dataset VQA 2.0 is open-sourced with a standard training and validation set. Based on the dataset, the fine-tuning details and instructions are clearly stated in the original work \cite{lxm, beit}, and model structures (\#layers, \#heads) are fixed. Our evaluation metrics are also standard and easy for implementation. 

\begin{table}[t!]
    \centering
    \begin{tabular}{l|l|l}
    \toprule
    \textbf{Parameter}   & \textbf{LXMERT}  & \textbf{BEiT3}                \\ \midrule
    Optimizer            & Adam         &    Adam      \\ 
    Epochs               & 4                    &   10        \\ 
    Learning rate        & \(5e^{-5}\)         &  \(3e^{-5}\)        \\ 
    Batch size           & 32                   &   16        \\ 
    Seeds                & 9595     & 42 \\ \bottomrule
    \end{tabular}
\caption{Model Parameters}
\label{tab3}
\end{table}
\end{document}